\documentclass{llncs}
\usepackage{makeidx}
\usepackage{times}
\usepackage{placeins}
\usepackage{latexsym}
\usepackage{hyperref}
\usepackage{graphicx}
\usepackage{multirow}
\usepackage{amssymb}
\usepackage{epigraph}
\begin{document}
%
%
%

%
%
\title{Semantic Relatedness for All (Languages): A Comparative Analysis of Multilingual Semantic Relatedness using
Machine Translation}
\titlerunning{Semantic Relatedness for All (Languages)}  
%

\author{Andr\'{e} Freitas\inst{1}, Siamak Barzegar\inst{2}, Juliano Efson Sales\inst{1},\\
Siegfried Handschuh\inst{1} and Brian Davis\inst{1}}
\authorrunning{Andr\'{e} Freitas et al.} 
\institute{Department of Computer Science and Mathematics - University of Passau\\
	    Innstrasse 43, ITZ-110, 94032 Passau, Germany\\
{\tt \{andre.freitas,juliano-sales,siegfried.handschuh\}@uni-passau.de}\\
\and
Insight Centre for Data Analytics - National University of Ireland, Galway\\
  	IDA Business Park, Lower Dangan, Galway, Ireland\\
  {\tt \{siamak.barzegar,brian.davis\}@insight-centre.org}}

\maketitle              

\begin{abstract}
This paper provides a comparative analysis of the performance of four state-of-the-art distributional semantic models (DSMs) over 11 languages, contrasting the native language-specific models with the use of machine translation over English-based DSMs. The experimental results show that there is a significant improvement (average of 16.7\% for the Spearman correlation) by using state-of-the-art machine translation approaches. The results also show that the benefit of using the most informative corpus outweighs the possible errors introduced by the machine translation. For all languages, the combination of machine translation over the Word2Vec English distributional model provided the best
results consistently (average Spearman correlation of 0.68).

\end{abstract}
\section{Introduction}

Distributional Semantic Models (DSM) are consolidating themselves as fundamental components for supporting automatic
semantic interpretation in different application scenarios in natural language processing. From \textit{question
answering systems}, to \textit{semantic search} and \textit{text entailment}, distributional semantic models support a
scalable approach for representing the meaning of words, which can automatically capture comprehensive
associative commonsense information by analysing word-context patterns in large-scale corpora in an unsupervised or
semi-supervised fashion\cite{thesisAndre,turney,linse}.

However, distributional semantic models are strongly dependent on the size and the quality of the reference corpora,
which embeds the commonsense knowledge necessary to build comprehensive models. While high-quality texts containing
large-scale commonsense information are present in English, such as Wikipedia, other languages may lack
sufficient textual support to build distributional models.

To address this problem, this paper investigates how different distributional semantic models built from
corpora in different languages and with different sizes perform in computing semantic relatedness similarity and
relatedness tasks. Additionally, we analyse the role of machine translation approaches to support the construction of better distributional vectors and for computing semantic similarity and relatedness measures for other languages.
In other words, in the case that there is not enough information to create a DSM for a particular language, this work aims at evaluating whether the benefit of corpora volume for English outperforms the error introduced by machine translation.

Given a pair of words and a human judgement score that represents the semantic relatedness of these two words, the
evaluation method aims at indicating how close distributional models score to humans. Three widely used word-pairs
datasets are employed in this work: Miller \& Charles (MC)\cite{miller1991contextual}, Rubenstein \& Goodenough
(RG)\cite{rubenstein1965contextual} and WordSimilarity 353 (WS-353)\cite{finkelstein2001placing}.

\begin{figure}[!htbp]
\centering
\includegraphics[width=0.8\textwidth]{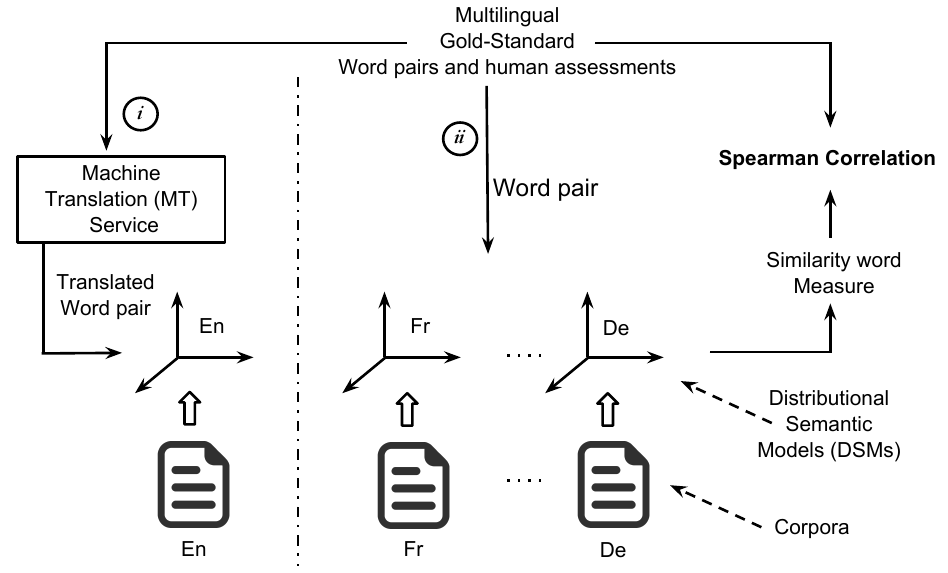}
\caption{Depiction of the experimental setup of the experiment.}
\label{fig:experimental_setup}
\end{figure}

In the proposed model the word-pairs datasets are translated into English as a reference language and the distributional
vectors are defined over the target end model (Figure \ref{fig:experimental_setup}). Despite the simplicity of the proposed method based on machine translation, there is a high relevance for the distributional semantics user/practitioner due to its simplicity of use and the significant improvement in the results.

This work presents a systematic study involving 11 languages and four distributional semantic models (DSMs), providing a
comparative quantitative analysis of the performance of the distributional models and the impact of machine
translation approaches for different models.

In summary, this paper answers the following research questions:

\begin{enumerate}

\item Does machine translation to English perform better than the word vectors in the original language (for which
languages and for which distributional semantic models)?

\item Which DSMs and languages benefit more and less from the translation?  

\item What is the quality of state-of-the-art machine translation approaches for word pairs (for each language)? 

\end{enumerate}

Moreover, this paper contributes with two resources which can be used by the community to evaluate multi-lingual
semantic similarity and relatedness models: \emph{(i)} a high quality manual translation of the three
word-pairs datasets - Miller \& Charles (MC)\cite{miller1991contextual}, Rubenstein \& Goodenough
(RG)\cite{rubenstein1965contextual} and WordSimilarity 353 (WS-353)\cite{finkelstein2001placing} - for 10 languages and
\emph{(ii)} the 44 pre-computed distributional models (four distributional models for each one of the 11 languages)
which can be accessed as a service\footnote{The service is available at \url{http://rebrand.ly/dinfra}.}, together with
the multi-lingual approaches mediated by machine translation.

This paper is organised as follows: Section \ref{related} describes the related work, Section \ref{setup} describes the
experimental setting; while Section \ref{results} analyses the results and provides the comparative analysis from
different models and languages, Finally, Section \ref{conclision} provides the conclusion.

\section{Related Work} \label{related}

Mostof related work has concentrated on leveraging joint multilingual information to improve the performance of the models.
 
Faruqui \& Dyer\cite{faruqui-dyer:2014:EACL} use the distributional invariance across languages and propose a technique based on
canonical correlation analysis (CCA) for merging multilingual evidence into vectors generated monolingually. They
evaluate the resulting word representations on semantic similarity/relatedness evaluation tasks, showing the improvement
of multi-lingual over the monolingual scenario.

Utt \& Pado\cite{utt-pado:2014:tacl}, develop methods that take advantage of the availability of annotated corpora in English using a translation-based approach to transport the word-link-word co-occurrences to support the creation of syntax-based DSMs.

Navigli \& Ponzetto\cite{navigli2012babelrelate} propose an approach to compute semantic relatedness exploiting the
joint contribution of different languages mediated by lexical and semantic knowledge bases. The proposed model uses a
graph-based approach of joint multi-lingual disambiguated senses which outperforms the monolingual scenario and
achieves competitive results for both resource-rich and resource-poor languages.

Zou et al.\cite{zou2013bilingual} describe an unsupervised semantic embedding (bilingual embedding) for words across two
languages that represent semantic information of monolingual words, but also semantic relationships across different
languages. The motivation of their works was based on the fact that it is hard to identify semantic similarities
across languages, specially when co-occurrences words are rare in the training parallel text. Al-Rfou et
al.\cite{al2013polyglot} produced multilingual word embeddings for about 100 languages using Wikipedia as the reference corpora.

Comparatively, this work aims at providing a comparative analysis of existing state-of-the-art distributional semantic models for different languages as well as analyzing the impact of a machine translation over an English DSM.

\section{Experimental Setup} \label{setup}

The experimental setup consists of the instantiation of four distributional semantic models (Explicit Semantic Analysis
(ESA)\cite{gabrilovich2007computing}, Latent Semantic Analysis (LSA)\cite{landauer1998introduction}, Word2Vec
(W2V)\cite{mikolov2013efficient} and Global Vectors (GloVe)\cite{pennington2014Glove}) in 11 different languages -
English, German, French, Italian, Spanish, Portuguese, Dutch, Russian, Swedish, Arabic and Farsi.

The DSMs were generated from Wikipedia dumps (January 2015), which were preprocessed by lowercasing, stemming and
removing stopwords. For LSA and ESA, the models were generated using the SSpace Package\cite{sspace}, while W2V and GloVe were
generated using the code shared by the respective authors. For the experiment the vector dimensions for LSA, W2V and
GloVe were set to 300 while ESA was defined with 1500 dimensions. The difference of size occurs because ESA is composed
of sparse vectors. All models used in the generation process the default parameters defined in each implementation.

Each distributional model was evaluated for the task of computing semantic similarity and relatedness measures using
three human-annotated gold standard datasets: Miller \& Charles (MC)\cite{miller1991contextual}, Rubenstein \&
Goodenough (RG)\cite{rubenstein1965contextual} and WordSimilarity 353 (WS-353)\cite{finkelstein2001placing}. As these
word-pairs datasets were originally in English, except for those language available in previous works
(\cite{faruqui2014community,camacho2015framework}), the word pairs were translated and reviewed with the help of
professional translators, skilled in data localisation tasks. The datasets are available at
\url{http://rebrand.ly/multilingual-pairs}.

Two automatic machine translation approaches were evaluated: the Google Translate Service and the Microsoft Bing
Translation Service. As Google Translate Service performed 16\% better for overall word-pairs translations, this was set
as the main machine translation model.

The DInfra platform \cite{barzegar2015dinfra} provided the DSMs used in the work. To support experimental
reproducibility, both experimental data and software are available at \url{http://rebrand.ly/dinfra}.

\section{Evaluation \& Results} \label{results}

\subsection{Spearman Correlation and Corpus Size}

Table \ref{tbl:correlation} shows the correlation between the average Spearman correlation values for each DSM and two
indicators of corpus size: \# of tokens and \# of unique tokens.

ESA is consistently more robust (on average) than the other models in relation to the corpus size due the
fact that ESA has larger context windows in opposition to the other distributional models. While ESA considers the whole
document as its context window, the other models are restricted to five (LSA) and ten (Word2Vec and GloVe) words. 

Another observation is that the evaluation of the WS-353 dataset is more dependent on the corpus size, which can be
explained by the broader number of semantic relations expressed under the semantic relatedness umbrella.

Table \ref{corpus_data} shows the size of each corpus in different languages regarding the number of unique tokens and
the number of tokens.

\begin{table*}[ht]
	\centering
	\begin{tabular}{|c|c|c|c|c|c|c|}
		\hline 
		
		\bf Gold standard & \multicolumn {2} {|c|}{ MC} &\multicolumn {2} {|c|}{ RG} & \multicolumn {2} {|c|}{ WS353} \\  \hline
		 & unique tokens & tokens & unique tokens & tokens & unique tokens & tokens \\ \hline
		ESA &  0.39 & \textit{0.48} & 0.67 & \bf 0.73 & \textit{0.33} & \textit{ 0.39} \\
		LSA & \bf 0.74 & \bf 0.75 & \bf 0.82 & 0.68 & \bf 0.64 & 0.66 \\
		W2V & 0.43 & 0.58 & 0.71 & 0.72 & 0.57 & \bf 0.79 \\
		Glove & \textit{0.34} & 0.51 & \textit{0.51} & \textit{0.61} & 0.59 & 0.63 \\
		\hline
	\end{tabular}
	\caption{\label{Table1} Correlation between corpus size and different models.}
	\label{tbl:correlation}
\end{table*}

\subsection{Word-pair Machine Translation Quality}

The second step evaluates the accuracy of state-of-the-art machine translation approaches for word-pairs (Table
3). The accuracy of the translation for the WS-353 word pairs significantly outperforms the other
datasets. This shows that the higher semantic distance between word pairs (semantic relatedness) has the benefit of
increasing the contextual information during the machine translation process, subsequently improving the mutual
disambiguation process.

\begin{table}[ht]
\centering
\begin{tabular}{|c|c|c|}
\hline
\bf \hspace{0.5cm}lang\hspace{0.5cm} & \bf \hspace{0.5cm}unique tokens\hspace{0.5cm} & \bf \hspace{0.5cm}tokens\hspace{0.5cm} \\ \hline
\bf en   & 4.238      & 902.044 \\ \hline
\bf de   & 4.233      & 312.380 \\ \hline
\bf fr   & 1.749      & 247.492 \\ \hline
\bf ru   & 1.766      & 202.163 \\ \hline
\bf it   & 1.411      & 178.378 \\ \hline
\bf nl   & 2.021      & 105.224 \\ \hline
\bf pt   & 0.873       & 96.712  \\ \hline
\bf sv   & 1.730      & 82.376  \\ \hline
\bf es   & 0.829       & 76.587  \\ \hline
\bf ar   & 1.653      & 46.481  \\ \hline
\bf fa   & 0.925       & 32.557  \\ \hline
\end{tabular}
\caption{The sizes of the corpora in terms of the number of unique tokens and tokens (scale of $10^6$).}
\label{corpus_data}
\end{table}

\begin{table}[ht]
\centering
\begin{tabular}{|c|c|c|c|c|c|c|c|c|c|c|}
\hline \bf dataset/lang & \bf	de	&	\bf	fr	&	\bf	ru	&	\bf	it	&	\bf	nl	&	\bf	pt	&	\bf	sv	&	\bf	es	&	\bf	ar	&	\bf	fa \\ \hline
MC	&		0.48	&		0.47	&		0.58	&		0.42	&		0.57	&		0.60	&		0.55	&		0.60	&		0.53	&		0.38 \\
RG	&		0.45	&		0.65	&		0.53	&		0.41	&		0.59	&		0.51	&		0.58	&		0.59	&		0.43	&		0.36 \\
WS353	&	\textbf{0.78}	&		\textbf{0.85}	&		\textbf{0.76}	&		\textbf{0.76}	&		\textbf{0.85}	&		\textbf{0.81}	&		\textbf{0.78}	&		\textbf{0.79}	&		\textbf{0.57}	&		\textbf{0.43} \\
\hline
\end{tabular}
\label{tbl:trans_acc}
\caption{Translation accuracy.}
\end{table}

\begin{table}[ht]
\small
\centering
\resizebox{\textwidth}{!}{\begin{tabular}{|c|c|c|c|c|c|c|c|c|c|c|c|c|c|c|}
\hline 
\bf DS	&	\bf Models	&	\bf en	&	\bf de	&	\bf fr	&	\bf ru	&	\bf it	&	\bf nl	&	\bf pt	&	\bf sv	&	\bf es	&	\bf ar	&	\bf fa	&	\bf Model AVG.	&	\bf DS AVG. \\ 
\hline 
\multirow{5}{*}{MC}	&	 ESA	&	 0.69	&	 0.67	&	 0.54	&	 0.66	&	 0.37	&	 0.54	&	 \textbf{0.67}	&	 0.37	&	 0.58	&	 0.37	&	 0.56	&	 0.53	&	 \textbf{0.56}  \\
&	 LSA	&	 0.79	&	 \textbf{0.70}	&	 0.55	&	 0.63	&	 0.58	&	 0.55	&	 0.41	&	 \textbf{0.58}	&	 0.66	&	 \textbf{0.46}	&	 0.45	&	 0.56	&	  \\
&	 W2V	&	\textbf{0.84}	&	\textbf{ 0.70}	&	 0.55	&	 0.64	&	 \textbf{0.74}	&	\textbf{ 0.57}	&	 0.37	&	 0.40	&	 \textbf{0.74}	&	 0.38	&	 \textbf{0.68}	&	\textbf{ 0.58}	&	 \\
&	 Glove	&	 0.69	&	 0.64	&	 \textbf{0.64}	&	 \textbf{0.76}	&	 0.51	&	 0.55	&	 0.62	&	 0.40	&	 0.65	&	 0.38	&	 0.45	&	 0.56	&	 \\
\hline 
\multirow{5}{*}{RG}	&	 ESA	&	 0.80	&	 0.68	&	 0.45	&	 0.63	&	 0.50	&	 0.58	&	 0.51	&	 0.50	&	 0.59	&	 0.36	&	 0.57	&	 0.54	&	 0.53 \\
&	 LSA	&	 0.72	&	 0.65	&	 0.30	&	 0.51	&	 0.48	&	 0.52	&	 0.30	&	 0.53	&	 0.35	&	 0.35	&	 0.46	&	 0.45	&	 \\
&	 W2V	&	 \textbf{0.85}	&	 \textbf{0.78}	&	 \textbf{0.57}	&	 0.64	&	 \textbf{0.69}	&	 \textbf{0.63}	&	 0.42	&	 \textbf{0.57}	&	 \textbf{0.64}	&	 \textbf{0.36}	&	 0.55	&	 \textbf{0.58}	&	 \\
&	 Glove	&	 0.74	&	 0.69	&	 0.50	&	\textbf{ 0.70}	&	 0.59	&	 0.54	&	 \textbf{0.52}	&	 0.49	&	 0.61	&	 0.32	&	 \textbf{0.59}	&	 0.56	&	 \\
\hline
\multirow{5}{*}{WS353}	&	 ESA	&	 0.50	&	 0.39	&	 0.32	&	 0.44	&	 0.34	&	 0.53	&	 0.44	&	 0.43	&	 0.37	&	 0.26	&	 0.37	&	 0.39	&	 0.41 \\
&	 LSA	&	 0.54	&	 0.45	&	 0.35	&	 0.40	&	 0.33	&	 0.47	&	 0.39	&	 0.40	&	 0.36	&	 0.28	&	 0.43	&	 0.39	&	 \\
&	 W2V	&	 \textbf{0.69}	&	 \textbf{0.54}	&	 \textbf{0.50}	&	 \textbf{0.53}	&	 \textbf{0.50}	&	 \textbf{0.58}	&	 \textbf{0.53}	&	 \textbf{0.45}	&	 \textbf{0.53}	&	 \textbf{0.44}	&	 \textbf{0.53}	&	 \textbf{0.51}	&	 \\
&	 Glove	&	 0.49	&	 0.41	&	 0.34	&	 0.42	&	 0.30	&	 0.46	&	 0.38	&	 0.33	&	 0.32	&	 0.26	&	 0.36	&	 0.36	&	 \\
\hline
&	 Lang AVG.	&	 0.70	&	 0.61	&	 0.47	&	 0.58	&	 0.49	&	 0.54	&	 0.46	&	 0.45	&	 0.53	&	 0.35	&	 0.50	&	 0.50	&	\\
\hline
\end{tabular}}
\label{tbl:language_specific}			
\caption{Spearman correlation for the language-specific models.}
\end{table}

\begin{table}[ht]
\small
\centering
\resizebox{\textwidth}{!}{\begin{tabular}{|c|c|c|c|c|c|c|c|c|c|c|c|c|c|}
\hline 
\bf DS	&	\bf Models	& \bf de	&	\bf fr	&	\bf ru	&	\bf it	&	\bf nl	&	\bf pt	&	\bf sv	&	\bf es	&	\bf ar	&	\bf fa	&	\bf Model AVG.	&	\bf Diff.  \\ 
\hline 
\multirow{5}{*}{MC}	&	 ESA-MT	&	 0.55	&	 0.53	&	 0.42	&	 0.38	&	 0.45	&	 0.38	&	 0.48	&	 0.39	&	 0.31	&	 0.58	&	 0.45	&	 -0.08 (-15.1\%) \\
&	 LSA-MT	&	 0.61	&	 0.72	&	 0.65	&	 0.67	&	 0.66	&	 0.70	&	 0.74	&	 0.78	&	 0.69	&	 0.75	&	 0.70	&	 0.14 (25.0\%)\\
&	 W2V-MT	&	 \textbf{0.68}	&	 \textbf{0.79}	&	 \textbf{0.68}	&	 \textbf{0.77}	&	 \textbf{0.69}	&	 \textbf{0.76}	&	 \textbf{0.81}	&	 \textbf{0.83}	&	 \textbf{0.71}	&	 0.74	&	 \textbf{0.75}	&	 \textbf{0.17 (29.3\%)} \\
&	 GloVe-MT	&	 0.45	&	 0.78	&	 0.67	&	 0.64	&	 0.63	&	 0.56	&	 0.61	&	 0.82	&	 0.69	&	 \textbf{0.79}	&	 0.66	&	 0.10 (17.9\%) \\
\hline
\multirow{5}{*}{RG}	&	 ESA-MT	&	 0.62	&	 0.53	&	 0.52	&	 0.61	&	 0.63	&	 0.57	&	 0.56	&	 0.47	&	 0.38	&	 0.71	&	 0.56	&	 0.02 (3.7\%) \\
&	 LSA-MT	&	 0.63	&	 0.62	&	 0.59	&	 0.74	&	 0.67	&	 0.64	&	 0.67	&	 0.62	&	 0.55	&	 0.70	&	 0.64	&	 \textbf{0.19 (42.2\%)} \\
&	 W2V-MT	&	 \textbf{0.69}	&	 \textbf{0.79}	&	 0.69	&	 \textbf{0.78}	&	 0.74	&	 \textbf{0.75}	&	 \textbf{0.71}	&	 \textbf{0.73}	&	 0.57	&	 0.79	&	 \textbf{0.72}	&	 0.14 (24.1\%) \\
&	 GloVe-MT	&	 0.62	&	 0.77	&	 \textbf{0.71}	&	 0.77	&	 \textbf{0.78}	&	 0.66	&	 0.66	&	 0.72	&	 \textbf{0.65}	&	 \textbf{0.80}	&	 0.71	&	 0.15 (26.8\%) \\
\hline
\multirow{5}{*}{WS353}	&	 ESA-MT	&	 0.42	&	 0.45	&	 0.41	&	 0.41	&	 0.44	&	 0.43	&	 0.40	&	 0.35	&	 0.42	&	 0.32	&	 0.40	&	 0.01 (2.6\%) \\
&	 LSA-MT	&	 0.51	&	 0.51	&	 0.47	&	 0.48	&	 0.51	&	 0.39	&	 0.51	&	 0.44	&	 0.37	&	 0.43	&	 0.46	&	 \textbf{0.07 (17.9\%)} \\
&	 W2V-MT	&	 \textbf{0.62}	&	 \textbf{0.59}	&	 \textbf{0.57}	&	 \textbf{0.57}	&	 \textbf{0.63}	&	 \textbf{0.51}	&	 \textbf{0.59}	&	 \textbf{0.55}	&	 \textbf{0.50}	&	 \textbf{0.52}	&	 \textbf{0.57}	&	 0.06 (11.8\%) \\
&	 GloVe-MT	&	 0.45	&	 0.48	&	 0.42	&	 0.43	&	 0.46	&	 0.33	&	 0.42	&	 0.41	&	 0.33	&	 0.37	&	 0.41	&	 0.05 (13.9\%) \\
\hline
&	Lang AVG.	&	 0.57	&	 0.63	&	 0.57	&	 0.60	&	 0.61	&	 0.56	&	 0.60	&	 0.59	&	 0.52	&	 0.63	&	 0.56	&	\\
\hline
\end{tabular}}
\label{tbl:machine_translation}
\caption{Spearman correlation for the machine translation models over the English corpora. \emph{Diff.} represents the difference of machine translation score minus the language specific.}
\end{table}

\begin{table}[ht]
\small
\centering
\begin{tabular}{|c|c|c|c|c|c|c|c|c|c|c|c|c|c|}
\hline 
\bf DS	&	\bf M	& \bf de	&	\bf fr	&	\bf ru	&	\bf it	&	\bf nl	&	\bf pt	&	\bf sv	&	\bf es	&	\bf ar	&	\bf fa	&	\bf M. AVG	&	\bf DS. AVG   \\ 
\hline 
\multirow{5}{*}{MC} & ESA & -0.18 & -0.03 & -0.36 & 0.03 & -0.16 & -0.44 & 0.31 & -0.32 & -0.16 & 0.03 & -0.13 & \multirow{5}{*}{\textbf{0.41}} \\
& LSA & -0.13 & 0.31 & 0.04 & 0.16 & 0.20 & 0.70 & 0.27 & 0.17 & 0.50 & 0.68 & 0.29 &  \\
& W2V & -0.02 & 0.43 & 0.07 & 0.05 & 0.21 & \textbf{1.04} & 1.00 & 0.13 & \textbf{0.88} & 0.09 & 0.39 & \\
& GloVe & -0.31 & 0.22 & -0.11 & 0.25 & 0.14 & -0.10 & 0.51 & \textbf{0.26} & 0.85 & 0.75 & 0.25 & \\
\hline
\multirow{5}{*}{RG} & ESA & -0.09 & 0.19 & -0.18 & 0.21 & 0.08 & 0.11 & 0.12 & -0.19 & 0.06 & 0.25 & 0.06 & \multirow{5}{*}{\textbf{0.41}} \\
& LSA & -0.03 & 1.04 & 0.14 & 0.52 & 0.30 & \textbf{1.15} & 0.26 & \textbf{0.77} & 0.57 & 0.52 & 0.52 &  \\
& W2V & -0.11 & 0.39 & 0.08 & 0.14 & 0.18 & 0.76 & 0.23 & 0.14 & 0.59 & 0.44 & 0.28 &  \\
& GloVe & -0.11 & 0.55 & 0.01 & 0.31 & \textbf{0.43} & 0.28 & 0.35 & 0.17 & \textbf{1.04} & 0.36 & 0.34 &  \\
\hline
\multirow{5}{*}{WS353} & ESA & 0.08 & 0.40 & -0.07 & 0.18 & -0.18 & -0.02 & -0.07 & -0.07 & 0.60 & -0.13 & 0.07 & \multirow{5}{*}{0.36} \\
& LSA & 0.12 & 0.43 & 0.19 & 0.45 & 0.09 & -0.01 & 0.27 & 0.21 & 0.34 & 0.01 & 0.21 &  \\
& W2V & 0.14 & 0.19 & 0.09 & 0.14 & 0.08 & -0.04 & 0.33 & 0.04 & 0.12 & 0.00 & 0.11 &  \\
& GloVe & 0.10 & 0.41 & 0.00 & 0.41 & 0.00 & -0.14 & 0.28 & 0.30 & 0.28 & 0.04 & 0.17 &  \\
\hline
& AVG & 0.06 & 0.52 & 0.13 & 0.36 & 0.23 & 0.29 & 0.70 & 0.22 & 0.59 & 0.82 &  & \\
\hline
\end{tabular}
\label{tbl:difference}
\caption{Difference between the language-specific and the machine translation approach. \textbf{M. AVG} represents the average of the models and \textbf{DS. AVG} represents the average of the datasets.}
\end{table}

For WS-353 the set of best-performing translations has an average accuracy of 80\% (with maximum 85\% and minimum
76\%). This value dropped significantly for Arabic and Farsi (average 50\%).

For MC and RG, the average translation accuracy for the semantic similarity pairs is 51.5\%. This difference may be a
result of a deficit of contextual information during the machine translation process. For these word-pairs datasets, the
difference between best translation performers and lower performers (across languages) is smaller. Additionally, the final translation accuracy for all languages and all word-pairs datasets is 59\%. French, Dutch and Spanish are the languages with best automatic translations.

\subsection{Language-Specific DSMs}

In the first part of the experiment, the Spearman correlations ($\rho$) between the human assessments and the computation
of the semantic similarity and relatedness for all DSMs instantiated for all languages were evaluated (Figure
\ref{fig:experimental_setup} \emph{(ii)}). Table \ref{tbl:language_specific} shows the Spearman correlation for each DSM
using language-specific corpora (without machine translation), for the three word-pairs datasets.

The comparative language-specific analysis indicates that English is the best-perfor-ming language (0.70), followed by
German (0.61). The lowest Spearman correlation was observed in Arabic (0.35). From the tested DSMs, W2V is consistently the
best-performing DSM (0.56). The language-specific DSMs achieved higher correlations for MC and RG (0.56 and 0.53,
respectively), in comparison to 0.41 for WS-353.

The results for the language-specific DSMs were contrasted to the machine translation (MT) approach, according to the
diagram depicted in Figure \ref{fig:experimental_setup} \emph{(i)}. The Spearman correlation for the MT-mediated approach are shown in
Table \ref{tbl:machine_translation}.

\subsection{Machine Translation based Semantic Relatedness}

Using the MT models, W2V is consistently the best performing DSM (average 0.68), while ESA is consistently the worst
performing model (0.47). We can interpret this result by stating that the benefit of using machine translation for ESA
does not introduces significant performance improvements in comparison to the language-specific baselines.

The best performing languages are French and Farsi ($\rho$ = 0.63). The Spearman correlation variance across languages
in the MT models is low, as the impact of the use of the English corpus on the DSM model has a higher positive impact on
the results in comparison to the variation of the quality of the machine translation. The results for all languages
achieve very similar correlation values.

The impact of the MT model can be better interpreted by examining the difference between the machine translation and
the domain-specific models (depicted in Table 6). LSA accounts for the largest average
percent improvement (28.4\%) using the MT model, while ESA accounts for the lowest value (-2.9\%). As previously
noticed, this can be explained by the sensitivity of these models to the corpus size due to the dimensional reduction
strategy (LSA) or the broader context window (ESA). The remaining models accounted for substantial improvements (W2V =
21.7\%, GloVe = 19.5\%).

Arabic and French achieved the highest percent gains (47\% and 38\%, respectively), while German accounts for worst
results (-4\%).These numbers are consistent with the corpus size. For German, the result shows that the corpus volume of
the German Wikipedia crossed a threshold size (34\% of the English corpus) above which improvements for computing
semantic similarity for the target word-pairs dataset might be marginally relevant, while the translation error accounts
negatively in the final result.

The average improvement for the MT over the language specific model for each word-pairs dataset is consistently
significant: MC = 20\%, RG = 30\% and WS\-353 = 14\%.


\subsection{Summary}
Below, the interpretation of the results are summarised as the core research questions which we aim to answer with this paper:
\\\\
\noindent \textbf{Question 1:} Does machine translation to English perform better than the word vectors in the original
language (for which languages and for which distributional semantic models)?

\noindent Machine translation to English consistently performs better for all languages, with the exception of
German, which presents equivalent results for the language-specific models. The MT approach provides an average
improvement of 16.7\% over language-specific distributional semantic models.
\\\\
\noindent \textbf{Question 2:} Which DSMs or MT-DSMs work best for the set of analysed languages?

\noindent W2V-MT consistently performs as the best model for all word-pairs datasets and languages, except German, in which the difference between MT-W2V and language-speci-fic W2V is not significant.
\\\\
\noindent \textbf{Question 3:} What is the quality of state-of-the-art machine translation approaches for word-pairs?

\noindent The average translation accuracy for all languages and all word-pairs datasets is 59\%. Translation quality varies according to the nature of the word-pair (better translations are provided for word pairs which are semantically related compared to semantically similar word pairs), reaching a maximum of 85\% and a minimum of 36\% across different languages.
\\\\
For the distributional semantics user/practitioner, as a general practice, we recommend using W2V built over an English corpus, supported by machine translation. Additionally, the accuracy of state-of-the-art machine translation approaches work better for translating semantically related word pairs (in contrast to semantically similar word pairs).

\section{Conclusion} \label{conclision}

This work provides a comparative analysis of the performance of four state-of-the-art distributional semantic models
over 11 languages, contrasting the native language-specific models with the use of machine translation over English-based
DSMs. The experimental results show that there is a significant improvement (average of 16.7\% for the Spearman
correlation) by using off-the-shelf machine translation approaches and that the benefit of using a more
informative (English) corpus outweighs the possible errors introduced by the machine translation approach. The average
accuracy of the machine translation approach is 59\%. Moreover, for all languages, W2V showed consistently
better results, while ESA showed to be more robust concerning lower corpora sizes. For all languages, the
combination of machine translation over the W2V English distributional model provided the best results consistently
(average Spearman correlation of 0.68).

Future work will focus on the analysis and translation of two other word-pairs datasets:
SimLex-999\cite{hill2015simlex999} and MEN-3000\cite{bruni}.

\section*{Acknowledgments}
This publication has emanated from research supported by the National Council for Scientific and Technological
Development, Brazil (CNPq) and by a research grant from Science Foundation Ireland (SFI) under Grant Number
SFI/12/RC/2289.

%
%
\bibliographystyle{splncs03}
\bibliography{biblio}
\end{document}